\documentclass[a4paper,twoside]{article}

\usepackage{times, mathptmx, calc, amstext}
\usepackage{amsthm}
\usepackage{multicol, pslatex, apalike, algorithm2e, cite, xcolor}
\usepackage[bottom]{footmisc}
\usepackage[switch]{lineno}
\usepackage[USenglish]{babel}
\usepackage[colorlinks,allcolors=blue]{hyperref}
\usepackage{amsmath, amsfonts, amssymb, latexsym, enumerate, mathrsfs, breqn, tabularx, multirow, lipsum, graphicx, blindtext, wrapfig, mathtools, colortbl, orcidlink, comment, caption, subcaption}

\usepackage[noend]{algpseudocode}
\usepackage[singlespacing]{setspace} 
\usepackage[colorlinks,allcolors=blue]{hyperref}
\enlargethispage{1in}
\definecolor{bubblegum}{rgb}{0.99, 0.76, 0.8}
\definecolor{lightgreen}{rgb}{0.56, 0.93, 0.56}
\definecolor{LightRed}{RGB}{252,160,140}
\definecolor{LightBlue}{RGB}{140,186,252}
\newcolumntype{a}{>{\columncolor{LightRed}}c}

\DeclarePairedDelimiter{\ceil}{\lceil}{\rceil}
\usepackage{SCITEPRESS}  
\usepackage[obeyFinal, textsize=tiny]{todonotes}

\begin{document}
\title{Unsupervised Representation Learning of Complex Time Series for Maneuverability State Identification in Smart Mobility}

\author{\authorname{Thabang LEBESE\orcidAuthor{0000-0003-3042-3111}}
\affiliation{Université Clermont Auvergne, Clermont Auvergne INP, CNRS, LIMOS, F-63000, Clermont-Ferrand, France}
\href{mailto:thabang.lebese@sigma-clermont.fr}{thabang.lebese@sigma-clermont.fr}}

\onecolumn \maketitle \normalsize \setcounter{footnote}{0} \vfill
\section{\uppercase{Research Problem}}
\label{sec:objectives}

Multivariate Time Series (MTS) data capture temporal behaviors to provide invaluable insights into various physical dynamic phenomena. In smart mobility, MTS plays a crucial role in providing temporal dynamics of behaviors such as maneuver patterns, enabling early detection of anomalous behaviors while facilitating pro-activity in Prognostics and Health Management (PHM). In this work, we aim to address challenges associated with modeling MTS data collected from a vehicle using sensors. Our goal is to investigate the effectiveness of two distinct unsupervised representation learning approaches in identifying maneuvering states in smart mobility. Specifically, we focus on some bivariate accelerations extracted from 2.5 years of driving, where the dataset is non-stationary, long, noisy, and completely unlabeled, making manual labeling impractical. The approaches of interest are Temporal Neighborhood Coding for Maneuvering (TNC4Maneuvering) and Decoupled Local and Global Representation learner for Maneuvering (DLG4Maneuvering).\\

The main advantage of these frameworks is that they capture transferable insights in a form of representations from the data that can be effectively applied in multiple subsequent tasks, such as time-series classification, clustering, and multi-linear regression, which are the quantitative measures and qualitative measures, including visualization of representations themselves and resulting reconstructed MTS, respectively. We compare their effectiveness, where possible, in order to gain insights into which approach is more effective in identifying maneuvering states in smart mobility.
\section{\uppercase{OUTLINE OF OBJECTIVES}}

Modern transportation is now equipped with more sensors than ever before, making the term "smart mobility" more fitting. This improves efficiency, security, and helps keep up with ever-changing environmental and government regulations, while at the same time assisting in facilitating pro-activity in Prognostics and Health Management (PHM). The sensors collect large amounts of data during operation time on multiple parts of the vehicle, including but not limited to engine performance, external conditions, and tire states. However, the sensory measurements are different and unique to each operation time, rendering unique behaviors for each of those times where the states are a function of operational time or mileage and are unique. For example, the Global Positioning System (GPS) collects geographical data, while sensors inside the odometer read mileage coverage. For this reason, the resulting collective sensory data is high frequency (up to fractions of a second), lengthy, noisy, non-linear, and impractical to label. Therefore, it is very challenging to relate underlying behaviors/states of one sensor to the other. This highlights the need for advanced representation learning methods, which can output a vectorial summary from multi-sensory inputs of variables over a specific time window as initially motivated by authors \cite{bengio2007scaling}. These resultant vectors are taken as descriptors of latent behaviors of the physical system, such as the accelerations that are derived from GPS sensor output.\\

Here, we focus on a bivariate acceleration MTS dataset extracted from 2.5 years of driving. The motivation behind using the two accelerations is that they easily provide primary description of different physical maneuvers of a vehicle. Secondly, it is easy to relate vehicle maneuvers to driving behavior because driving generally involves three main actions: controlling the steering wheel, stepping on the accelerator, and pressing the brake pedal. These actions are captured by the two accelerations, namely the lateral acceleration $(a_{lat})$ and longitudinal acceleration $(a_{lon})$, which pertain to steering actions, accelerator, and brake pedal usage experienced by a vehicle, respectively. Hence, extracting representations of the accelerations can lead to improved performances of subsequent Machine Learning (ML)tasks that rely heavily on the quality of the representations.\\

Following our prior works \cite{tlebese} that focused on representation learning using simulated datasets, we further investigate and extend the effectiveness of the two distinct unsupervised representation learning approaches, Temporal Neighborhood Coding for Maneuvering (TNC4Maneuvering) and Decoupled Local and Global Representation learner for Maneuvering (DLG4Maneuvering), in identifying maneuvering states in smart mobility using a bivariate dataset. The methods are unsupervised and perform representation learning useful for extracting driving ``states'' to understand maneuverability in smart mobility in complex MTS. Our results demonstrate the potential of usability on downstream tasks and their robustness in identifying and locating temporal transitions between states without any prior knowledge about labels while improving the quality and interpretability of the identified underlying representations. Secondly, we also aim to compare their effectiveness by evaluating the two frameworks in various downstream tasks, including the quantitative measures such as the time-series classification, clustering, and multi-linear regression, where extracted representations are inputs, and qualitative indicators, which are visualizations of representations themselves and resulting reconstructed MTS, respectively. We use these indicators to check for visual interpretability of representations and thirdly the reconstruction of the DLG4Maneuvering in order to gain insights into which approach is more effective in identifying maneuver states in smart mobility.
\section{\uppercase{State of the Art}}

According to \cite{bengio2007scaling}, the quality of data representations is critical for performance of most Machine Learning (ML) models. This is especially true for complex data types such as time series, which can be high-dimensional, high-frequency, and non-stationary \cite{yang200610, langkvist2014review}. Due to the difficulty and impracticality of manually labeling time-series data, different ML methods are preferred, ranging from supervised, unsupervised, and semi-supervised approaches.\\

\textbf{ML for Vehicle Maneuvering:} Prior works have utilized statistical methods \cite{liu2022statistical, fadhloun2015vehicle, haas2004use, maurya2012study, hashimoto2022study} and GG-analysis\footnote{\href{https://optimumg.com/characterising-tracks-for-set-up-solutions/}{GG-analysis}} to extract representations of driver behavior. Traditional ML techniques like SVM, RF, NB, KNN, and MLP have been employed 
\cite{ouyang2019ensemble, zheng2017lane, ouyang2017improved, carlos2019smartphone}. \cite{haque2022driving} proposed a rule-based machine learning technique using a sequential covering algorithm for classifying driving maneuvers. However, these methods are limited in handling high-dimensional data with complex patterns, resulting in inferior performance compared to deep learning approaches.\\ 

\textbf{Unsupervised representation learning:} Although unsupervised representation learning has shown great success in various MTS tasks, its application to smart transportation MTS datasets is generally limited. Existing attempts, such as the application of Bag of Words (BoW) model in \cite{carlos2019smartphone}, led to a representation-like output with a focus on classifying aggressive driving maneuvers only. Such approaches do not generalize well, making them incapable of other alternative subsequent tasks.\\

Recent works explore contrastive learning for representation learning by contrasting similar and dissimilar instances. Examples include \cite{tonekaboni2021unsupervised,franceschi2019unsupervised, oord2018representation, lai2019contrastive, zerveas2021transformer, ijcai2021-324, yue2022ts2vec, hyvarinen2016unsupervised, eldele2022self}. Notable exceptions are \cite{woo2022cost}, which disentangles seasonal-trend features using time and frequency domains, and \cite{choi2023multi}, which jointly learns contextual, temporal, and transformation consistencies, later applying them to classification, forecasting, and anomaly detection tasks. To the best of our knowledge, our work is the first to utilize pure unsupervised representation learning of acceleration MTS, specifically for understanding vehicle maneuvering with capabilities to multitask downstream.\\

\textbf{Unsupervised generative modeling:} Recently, methods including Variational Auto-Encoder (VAE) approaches, have been limited in applications of smart transportation or automotive Multivariate Time-series (MTS) datasets. Existing methods like those by \cite{shouno2018deep}, \cite{bao2021prediction}, and \cite{arbabi2022learning} use VAE-based models to disentangle dynamic and static factors in driving maneuvers, their focus is primarily on clustering or predicting behaviors. However, these methods lack attention to the quality and interpretability of the representations. on the other hand, methods like DSVAE \cite{yingzhen2018disentangled}, S3VAE \cite{zhu2020s3vae}, and C-DSVAE \cite{bai2021contrastively} offer interesting approaches by emphasizing the generation of samples rather than just disentangling dynamic and static factors but they have not been explored in the context of smart transportation datasets.\\

A different line of work is aimed at disentangling global and local representations, although it is often applied to visual data for factorizing label-related variation. \cite{mathieu2016disentangling} and \cite{ma2020decoupling} used conditional generative models and empirical characteristics of VAE and flow models, respectively. However, these efforts were primarily tailored for specific downstream tasks. For instance, \cite{sen2019think}, \cite{wang2019deep}, and \cite{nguyen2021temporal} focused on improving forecasting using global and local patterns. Unlike most prior works that prioritize sample generation, our work uniquely emphasizes the quality and interpretability of representations. To the best of our knowledge, we once again can claim our study is the first to apply pure unsupervised generative modeling to acceleration MTS for understanding vehicle maneuvering and multitasking downstream.
\section{\uppercase{Methodology}}

\textbf{Notation:} Let $X\in \mathbb{R}^{F\times T}$  be a MTS sample with $F$ features and $T$ measurements. Each feature is derived from two latent variables: $z_g$ (global) and $Z_l$ (local). The global representation $z_g^{(i)}\in \mathbb{R}^{d_g}$ captures overall sample properties, and the local representation 
$Z_l^{(i)}\in \mathbb{R}^{d_l}$ is a set of vectors extracted from non-overlapping time series windows $W_t^{(i)}$ of size $\delta$. Each $Z_l^{(i)}$ encodes information from all features within a window. The overall MTS is divided into $W_t = \ceil{\frac{T}{t}},~t = 1,2, \dots$, in DLG4Maneuvering, whereas in TNC4Maneuvering we further subdivide each $t$ to get $W_t = \ceil{t-\frac{\delta}{2}, t+\frac{\delta}{2}},~\delta = 1,2, \dots$, of non-overlapping windows respectively. Global and local representation sizes are defined as $d_g = m$ and $d_l = M$, respectively, where $d_l = M<<F \times W_t$ and $d_g = m \leq M$. In the case of missing measurements, each sample includes a mask channel $M_{usk}^{(i)}\in \mathbb{R}^{F\times W_t}$ to indicate observed and missing points, allowing conversion of irregular MTS into regularly-sampled signals. Some of the differences is that in TNC4Maneuvering, there is no masking component, instead missing values are padded with zeros and there is no global encoder. Figure~\ref{fig:tnc-dgl} depicts how these notations holds in the proposed methods for representation learning and generative modeling frameworks for maneuverability extraction in smart transportation.

\subsection{Representation Learning} \label{sec:encoder_only}

\textbf{TNC4Maneuvering:} the backbone of our method is a non-linear composition function encoder $(Enc)$, typically a deep neural network, taking a static window $W_{t}$ centered at time $t$ with sub-length $\delta$ and $F$ number of features. A tuple of samples, an anchor $(W_t)$, a positive $(W_l)$ and negative $(W_k)$ windows are sampled from input MTS where each window
$W_{t,l,k}$ generates a representation vector $Z_{t,l,k} \in \mathbb{R}^M$, where $M << F \times W_{t,l,k}$ is the size of the vector. $W_l \text{ and } W_t \in N_t$ share the same neighborhood centered at $t$, while $W_k \in \overline{N}_t$ is at a distant non-neighborhood. The semantic similarities and dis-similarities between windows is controlled by the temporal neighborhood around $W_t$. This region is defined as a region where acceleration signals are relatively stationary compared to their pre and post-windows, they are therefore assumed to be generated from the same underlying maneuvering state. The objective function (\ref{eq:tnc_loss}) is a partial contrastive loss that learns signals via encoding and evaluates them using a Discriminator $(D)$ that identifies representations with similar underlying maneuverings.

{\footnotesize
\begin{dmath}\label{eq:tnc_loss}
  \mathcal{L} = - \mathbb{E}_{W_t \sim X} \Bigg[ 
 {\mathbb{E}_{W_l \sim N_t} \bigg[ \log \big(\mathcal{D}(Z_t,Z_l)\big) \bigg]} + {\mathbb{E}_{W_k \sim \overline{N}_t} \bigg[ w_t \log \big(\mathcal{D}(Z_t,Z_k)\big) + (1-w_t) \log \big(1- \mathcal{D}(Z_t,Z_k)\big) \bigg]} \Bigg].
\end{dmath}}

\begin{figure*}
 \begin{subfigure}{0.45\textwidth}
\includegraphics[width=\linewidth]{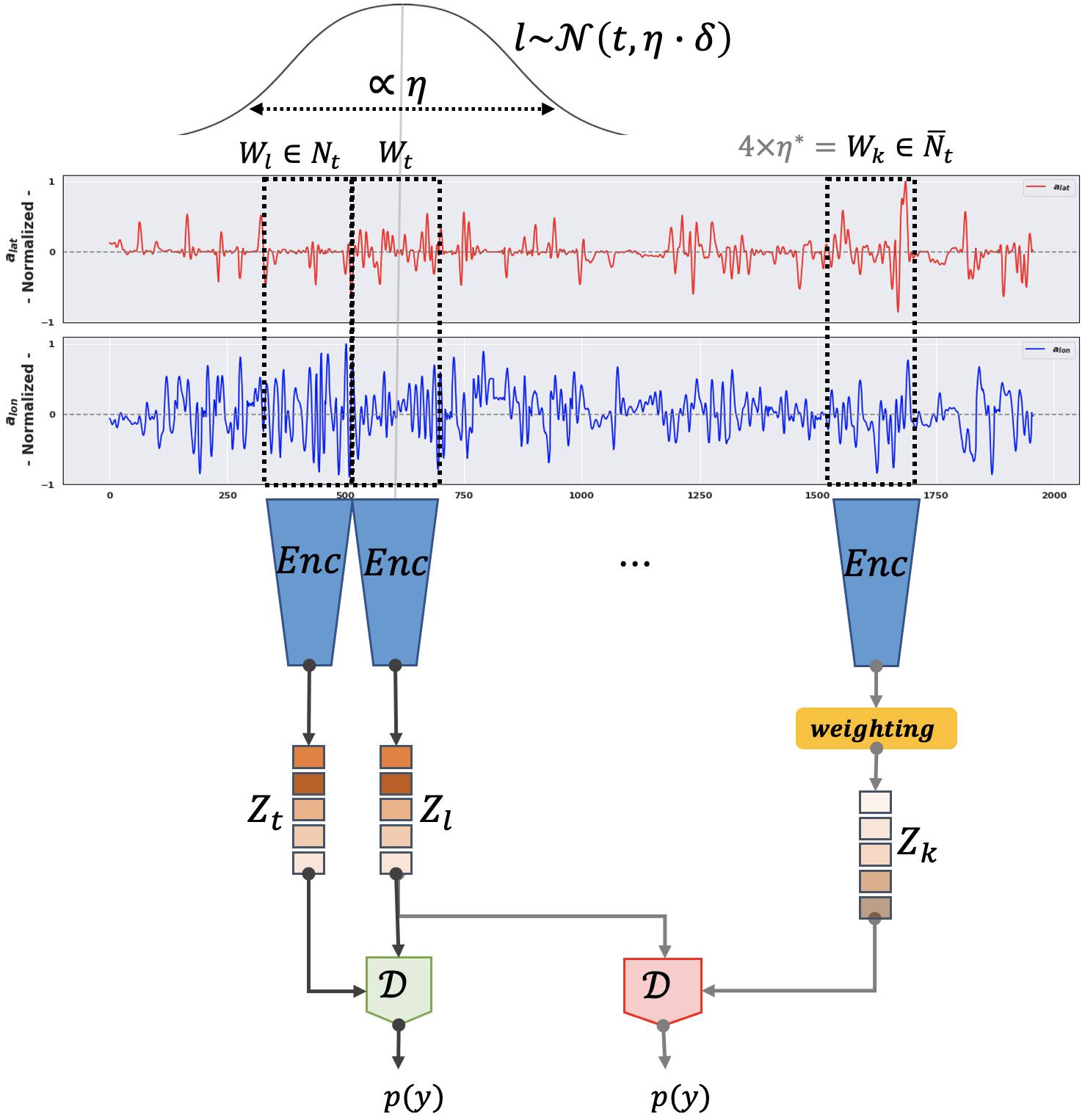}
\caption{\textbf{TNC4Maneuvering:} Encoder: $Enc(W_t)$, outputs representations $Z_t \in \mathbb R^{M}$, with Discriminator: $\mathcal{D}(Z_t, Z_{l\vee k}) \in [0,1]$.} \label{fig:tnc}
 \end{subfigure}%
 \hfill
 \begin{subfigure}{0.5\textwidth}
\includegraphics[width=\linewidth]{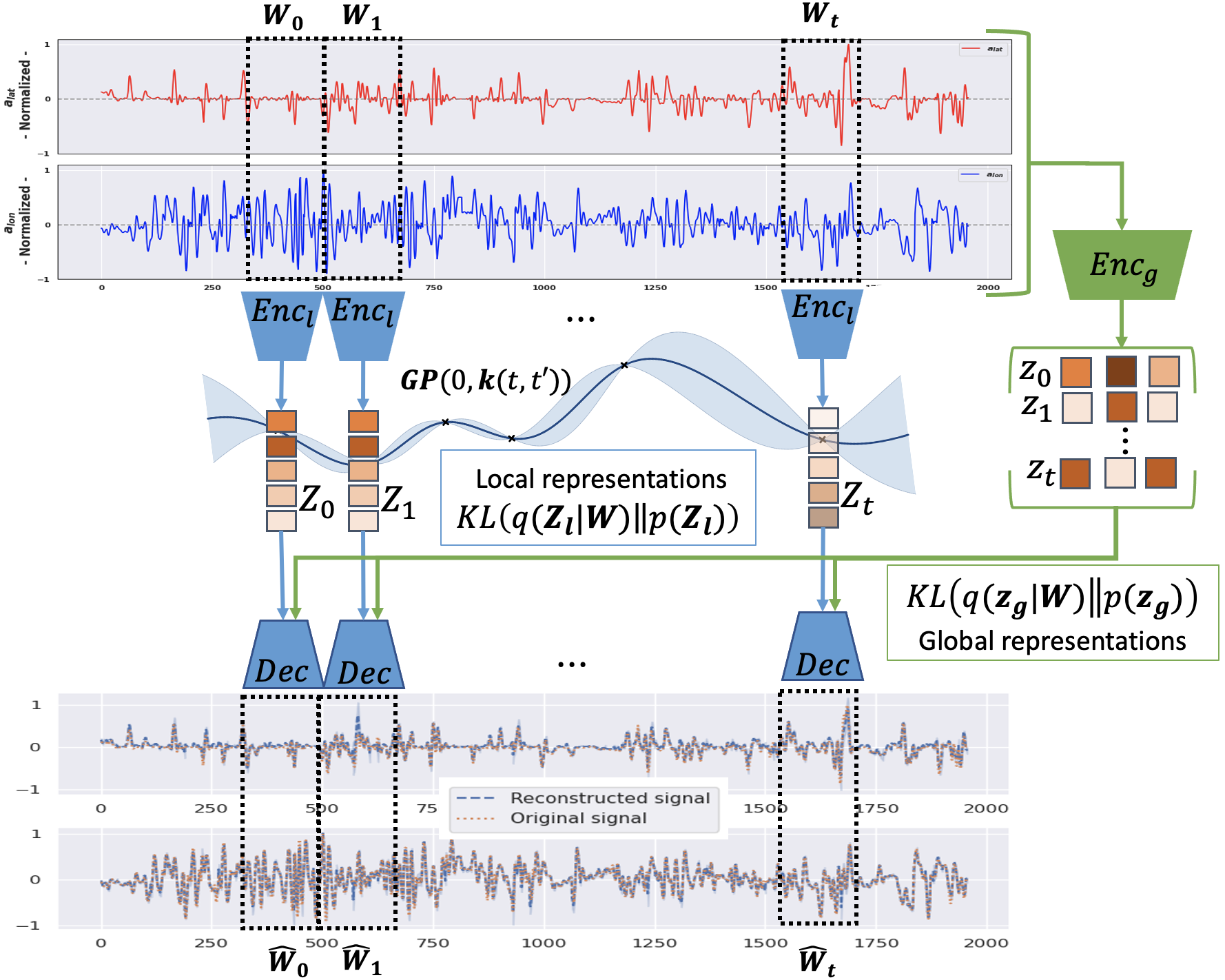}
\caption{\textbf{DLG4Maneuvering:} Local Encoder:
 $Enc_l(W_t)$, outputs local representations $Z_l \in \mathbb R^{M}$; global Encoder:
 $Enc_g(W_t)$, outputs global representations $z_g \in \mathbb R^{m}$ with priors $p(Z_l)$ being a zero-mean GP $(GP(0, k(t, t^{\prime}))$ and $Dec(Z_l, z_g)$ generates corresponding windows $\widehat{W}_t$ using from both representations.}
 \label{fig:dgl}
 \end{subfigure}
 \vskip 0.1in
 \caption{Representation learning and generative modeling frameworks for maneuverability extraction in smart transportation.\label{fig:tnc-dgl}}
\end{figure*}

The unit root test, Augmented Dickey-Fuller (ADF)\footnote{\href{https://github.com/bashtage/arch/}{arch.unitroot.ADF}} is used for determining relative stationarity region. Furthermore, the loss is weighted using ideas from Positive-Unlabeled (PU) learning to counter the potential sampling bias in the contrastive objective. This compensates for negative samples drawn from outside of the neighborhood which may in fact be similar to those of an anchor window. The overall framework is depicted in Figure~\ref{fig:tnc} and further details on this framework can be found in \cite{tonekaboni2021unsupervised, tlebese}.

\subsection{Generative Modeling} \label{sec:dlgreps}

\textbf{DLG4Maneuvering:} incorporates multiple components such as non-linear composition functions including local and global encoders $(Enc_l, Enc_g)$, and a decoder $(Dec_{g+l})$ implemented as deep neural networks. Due to absence of labels, counterfactual regularization is employed to enhance the informativeness of global representations. Learning follows a probabilistic generation, where the conditional likelihood distribution of the data is modeled as $W_t \sim p(W_t | Z_l, z_g)$, with $W_t$ associated with dynamic local representation $Z_t$. Priors for local representations use a Gaussian Process ($GP$) with dependencies represented as $GP(m(t), k(t, t'))$, where $m(t)$ is mean and $k(t, t')$ is the covariance function. The global representation $z_g$ remains constant within a window and follows a Gaussian distribution $\mathcal{N}(0, 1)$.\\

DLG4Maneuvering employs a variational approximation model, addressing three distributions: 1) The conditional likelihood distribution of MTS $p(W_t|Z_l, z_g)$ approximated by the decoder model $Dec(Z_l, z_g)$. 2) The posterior distribution over local representations $q(Z_l|W_t)$ approximated by the local encoder $Enc_l(W_t)$. 3) The posterior distribution of global representations $q(z_g|W_t)$ approximated by the global encoder $Enc_g(W_t)$, learning the parameters of the conditional distribution. To capture temporal dependencies between local representations, different Gaussian Process kernels are used. Each dimension of local representations (denoted as $j \in M$) is independently modeled over time, allowing for the capture of unique temporal behaviors characterized by distinct covariance structures. The objective function is an ELBO-based VAE~\cite{kingma2013auto} loss, given as:

{\footnotesize
\begin{dmath}\label{eq:DLGReps_loss}
  \mathcal{L} = \mathbb{E}_{Z_l, z_g} \Big[log(p(W_t|Z_l, z_g)) \Big] - \Big[D_{KL}(q(Z_l|W_t) || p(Z_l)) + D_{KL}(q(z_g|W_t) || p(z_g)) \Big]. 
\end{dmath}}

The negative reconstruction error given by the first term in this context is proportional to the mean squared error (MSE) of the input $W_t$ and the reconstruction given the probabilistic encoder and decoder when summed over a batch of samples with $\log(\cdot)$ ensuring realistic signal generation. Whereas the second terms $-[D_{KL}(\cdot||\cdot) + D_{KL}(\cdot||\cdot)]$ minimizes the distance between estimated distributions and their priors and are obtainable analytically. Authors \cite{burgess2018understanding} introduced scalar $\beta >1$ for weighting the KL-divergence with the goal of further disentangling the latent space. For such a $\beta$, each dimension is more closely related to features of the output, resulting in the so-called $\beta$-VAE method. Whereas \cite{otten2021event} introduced a B-VAE method by introducing a parameter $B \ll 1$ instead to emphasize for a better improvement of reconstructions. However, this also implies that $-[D_{KL}(\cdot||\cdot) + D_{KL}(\cdot||\cdot)]$ terms will be least important since there is a smaller penalty when the latent representation distributions are deviant from a standard Gaussian. The final objective function of equation~\eqref{eq:DLGReps_loss} is now written as,

{\footnotesize
\begin{dmath}\label{eq:DLGReps_Bloss}
  \mathcal{L} = \frac{1}{\kappa}\sum_{i=1}^{\kappa}(1-B)\cdot \text{MSE} + B [D_{KL}(\cdot||\cdot) + D_{KL}(\cdot||\cdot)]. 
\end{dmath}}

Where, $\kappa$ is the batch-size and the terms MSE and the two $D_{KL}$ are equivalent to the first and second terms in equation~\eqref{eq:DLGReps_loss}. A special case for $B = 0$ which is equivalent to a standard Auto-Encoder (AE) can be obtained. Priors for global representations $p(z_g)$ are assumed to be a standard Gaussian $\mathcal{N}(0,1)$, while priors for local representations $p(Z_l)$ use a zero-mean GP with different kernels and parameters to capture variances in dynamics at various time scales. Negative log-likelihoods are only estimated for observed measurements to account for missing values. $Enc_l$ has a higher encoding capacity and is prone to dominate information flow, potentially rendering $z_g$ as random noise, neglectable by the Decoder $D(Z_l, z_g)$.\\ 

To address this, a counterfactual regularization term $L_{reg}$ is introduced in the objective as a third term in equation~\eqref{eq:DLGReps_Bloss}. This term encourages $z_g$ to be informative while promoting disentanglement. In the training phase, each window sample $W_t^{(i)}$ is paired with a counterfactual sample $W^{*}$ generated without global properties. As the two representations ($z_g$ and $Z_l$) are independent, $Z_l^{(i)}$ cannot contain any information about $z_g^{(i)}$. Consequently, $z_g^{(i)}$ will have a low likelihood under the estimated posterior distribution $q(z_g|W^*)$. Utilizing the global encoder to estimate this posterior encourages a low likelihood ratio for $z_g$ to $z_g^*$. Hence, the counterfactual regularization promotes implicit independence between global and local variables given by an additional term in the objective as $L_{reg} = E_{z_g}, Z_l \frac{q(z_g | W_t^*)}{q(z_g^*|W_t^*)}$, where $\lambda$ is a counterfactual regularization weight making the final objective to be given as equation~\eqref{eq:DLGReps_Bloss_reg}. The overall overview depiction of DLG4Maneuvering framework is given in Figure~\ref{fig:dgl}.

{\footnotesize
\begin{dmath}\label{eq:DLGReps_Bloss_reg}
  \mathcal{L} = \frac{1}{\kappa}\sum_{i=1}^{\kappa}(1-B)\cdot \text{MSE} + B [D_{KL}(\cdot||\cdot) + D_{KL}(\cdot||\cdot)] + \lambda \cdot L_{reg}.
\end{dmath}}

\subsection{Model Details}\label{subsec:enoder}

From \cite{tlebese}, we replace the Bidirectional Recurrent Neural Network (BiRNN) \cite{schuster1997bidirectional} with an exponentially dilated Convolutional Neural Network (CNN) \cite{yu2015multi} with causality as backbone encoders for both models. The main reasons behind our setup is that in vanilla CNNs, the size of the receptive field can be linearly related to number of layers and the kernel width, but to cover longer temporal dependencies, larger receptive fields are required. But larger receptive fields require increase in number of layers making training process more difficult and both time and resource expensive. On the other hand, Recurrent Neural Networks (RNNs) and its special kinds such as Long Short Term Memory (LSTM) suffer from vanishing gradients and have trouble learning long temporal dependencies because the added memory retention components still have trouble learning very long-distance relationships due to the need of increased back-propagation steps needed for longer temporal dependencies. Hence in exponentially dilated convolutions can efficiently capture long-range dependencies without increasing network depth. They are a better option because, they enable increased receptive fields exponentially without loss in coverage with short-distance gradient propagation. Hence, for this reason they are a better option in such applications where integrating knowledge of wider context with less cost is crucial.\\

Our exponentially dilated CNN encoders are tailored for encoding MTS data into a lower-dimensional vector space, particularly suited for datasets with extended temporal dependencies and characteristics such as being non-Gaussian, intermittency, non-periodicity, and so on. Each encoder $Enc$, $Enc_l$ and $Enc_g$ comprises of three stacked convolutional layers, each using dilated convolutions to extract inter-temporal features. The dilation parameter exponentially increases ($2^i$ for the $i$-th layer), while fixed-size filters ($f \in \mathbb{N}$) preserve temporal resolution and alignment. The output undergoes global max pooling, compressing temporal information into a fixed-size vector. This result is flattened and processed by a linear layer, further reducing the dimensionality to produce an encoding of sizes $M$ and $m$, serving as compressed representations based from a window size $W_t$ respectively.\\

\textbf{Encoders:} The encoder designs in both TNC4Maneuvering and DLG4Maneuvering offer some level of flexibility by allowing customized encoder sizes ($M, m$), incorporating a classification component for compatibility with subsequent tasks in TNC4Maneuvering. The design choices presents several advantages, including enhanced generalization for downstream tasks and ease of adjusting encoder sizes ($M, m$). Each exponentially dilated convolution layer encodes data through a convolution operation with dilation defined by:

\begin{equation}
 F(s) = (W_t\star_{d}f)(s) = \sum_{i=0}^{k-1} f(i)W_{t}^{s-d \cdot i},
\end{equation}

where $F(s)$ represents the computed output on each layer for samples $s \in W_t$ ($\in \mathbb{R}^{F\times \delta}$), with a dilation rate of $d$, filter size $k$, and $(s-d \cdot i)$ accounting for the historical direction. We perform an 80/20 train/test data split with training epochs limited to 30 for both TNC4Maneuvering and DLG4Maneuvering, respectively. \\ 

\textbf{Decoder:} Our decoder $Dec(Z_l, z_g)$, exclusive to DLG4Maneuvering, generates corresponding windows $\widehat{W}_t$ using both representations. In this case we employ a vanilla RNN architecture, leveraging the distinctive capability of DLG4Maneuvering framework to separate global and local representations. Another advantage of the decoupled local and global representations is that they have already been condensed into representations of fewer dimensions. This reduction is particularly notable due to the uniqueness of the global representation across samples, justifying the use of a simple generation framework like an RNN.

\subsection{Hyper-parameter Selection} \label{sec:params}
For a meaningful comparison of these two distinct methods, we strive to align their hyperparameters wherever feasible. The hyperparameters for both TNC4Maneuvering and DLG4Maneuvering are provided in Table~\ref{table:DLGReps4Mparams}. While there is a potential for further tuning, particularly in the window size ($W_t$) and latent space dimension ($M$) as highlighted in \cite{tlebese}, here we have chosen to omit these adjustments in light of the main objectives of our study. We also take note of the oversight that \cite{tonekaboni2022decoupling} used a $\beta < 1$ in their objective function instead of $\beta > 1$ to conform with original works of \cite{burgess2018understanding}.

\begin{table}[ht]
\footnotesize
\caption{Selected hyper-parameters for training DLG4Maneuvering and TNC4Maneuvering.\label{table:DLGReps4Mparams}}\centering
\begin{tabular}{c|c|c} 
\hline
Parameter & DLG4Maneuvering & TNC4Maneuvering \\
\hline
$W_t$ &19 &19 \\
$w_t$ &- &0.05 \\
$\lambda$ &0.8 &- \\
B &0.01 &- \\ 
$M$ &16 &16 \\
$m$ &2 &- \\
$lr$ &0.001 &0.001 \\
Opt. &Adam &Adam \\
ADF &- &0.01 \\
Prior &RBF, Matern32 &- \\
Prior Scale &2, 1, 0.5, 0.25& - \\
Batch-size ($\kappa$)  &5 &5 \\
\hline
\end{tabular}
\end{table}

\subsection{Evaluation}
In order to evaluate the performance of TNC4Maneuvering and DLG4Maneuvering, we evaluate four downstream tasks namely, time-series classification, clustering, multi-linear regression and MTS Reconstruction which only applies for DLG4Maneuvering.\\

\begin{table*}[t]
\caption{Performances across multiple downstream tasks for TNC4Maneuvering and DLG4Maneuvering. \label{table:states}}
\centering
\resizebox{\textwidth}{!}{%
\begin{tabular}{|c|c|cc cc cc|} 
             \hline 
            \multicolumn{2}{|c}{} &
            \multicolumn{2}{c}{Classification} &
            \multicolumn{2}{c}{Clustering} &
            \multicolumn{2}{c|}{Regression} \\
            \hline
            Model &\rotatebox[origin=c]{0}{$W_t$} &
            \rotatebox[origin=c]{0}{AUPCR}&\rotatebox[origin=c]{0}{Accuracy}&
            \rotatebox[origin=c]{0}{Silhouette}&\rotatebox[origin=c]{0}{DBI}& \rotatebox[origin=c]{0}{$R^{2}$} & \rotatebox[origin=c]{0}{Loss}\\
            \hline
        \multirow{1}{*}[0em]{\rotatebox[origin=c]{0}{TNC4Maneuvering}} &19& 0.529 & 53.310 &0.273 &1.217 &-0.343 &0.449\\ 
        \multirow{1}{*}[0em]{\rotatebox[origin=c]{0}{DLG4Maneuvering}} &19 &0.998 &99.70 &0.143 &0.983 &-0.062 &0.355\\ 
        \hline 
        \end{tabular}}
\end{table*}


\textbf{Classification:} In this subsequent task, we employ a linear classifier due to its effectiveness in separating representations in high dimensions, assuming well-separated representations. In the TNC4Maneuvering model, setting the parameter $(classify = True)$ triggers the classification task, whereas in DLG4Maneuvering global representations $(m)$ are used as labels. The encoding are input to a classifier comprising a dropout layer to prevent overfitting and a linear layer mapping the encoding to predefined maneuver output classes ($n_{\text{classes}}$) for classification. We evaluate using prediction accuracy and the area under the precision-recall curve (AUPRC) score, specifically suitable for imbalanced classification settings. The classification algorithm learns relationships between representations and predefined maneuver labels (defined in section~\ref{subsec:car-data}), facilitating accurate prediction and categorization of maneuvering states.\\

\textbf{Clustering:} Clustering of representations assesses their separability in the latent space using k-means \cite{macqueen1967classification}, offering insights about resulting encoding properties with predefined maneuver labels (defined in section~\ref{subsec:car-data}). We employ two metrics for evaluation: the Silhouette score and Davies-Bouldin Index (DBI). The Silhouette score measures the similarity of an encoding within its assigned cluster versus adjacent clusters, ranging from $[-1,1]$. A higher score implies better cohesion. The DBI assesses both intra-cluster coherence and inter-cluster separation, with a lower score indicating better clusterability. Identified clusters in clustered representations are expected to reflect similar characteristics related to vehicle maneuver behavior.\\

\textbf{Regression:} In this subsequent task, peaks and valleys also known as turning points are collected. By taking consecutive differences between turning points and their square sums, quantifies their magnitudes in each window. This results to a vector $X_{man} \in \mathbb{R}^{M \times 1}$ as a summary. On the other hand, the resultant vector should offer insights into the intensity and characteristics of extrema fluctuations found in the datasets. We assume a linear mapping as a first trial where a vector $X_{man} \in \mathbb{R}^{M \times 1}$ is regressed by multivariate representations $Z \in \mathbb{R}^{M}$, although our perspective would be to propose a non-linear one. A train-test $(70/ 30)$ data split is performed, as evaluation coefficient of determination ($R^2$) and learning loss are used.\\

\textbf{Representations:} Visualized representations against acceleration signals over time enhances the understanding and interpretation of extracted maneuver state and how they are modeling in the latent space ($Z \in \mathbb{R}^{M}$). This visual metric is crucial for comprehending vehicle maneuvering as it provides insights into maneuver behavior through visualization, facilitating the recognition of changes in maneuver states over time. Capturing these changes clearly enables deeper insights into the severity or gentleness of driver maneuvers. \\

\textbf{Reconstructions:} DLG4Maneuvering disentangles global and local representations to enhance downstream modeling tasks. The interpretability of both representations can be directly linked to the quality of reconstructed samples. Assessing the quality of reconstructions is particularly valuable, as it serves as a meaningful metric linked to various subsequent tasks, such as forecasting, even in the presence of missing data. This visual metric proves crucial for understanding vehicle maneuverability, facilitating comprehension of how easily driving behavior can be reconstructed over time, irrespective of driving complexities. This approach further enables clearer understanding of the severity or gentleness of driver maneuvers. Our reliance on the method's ability to produce perfect reconstruction samples from interpretable local and global representations is a critical evaluation criterion.
\section{\uppercase{Intermediate results}}
\subsection{Acceleration Dataset} \label{subsec:car-data}

\begin{figure*}[t]
  \begin{subfigure}{0.49\textwidth}
\includegraphics[width=\linewidth]{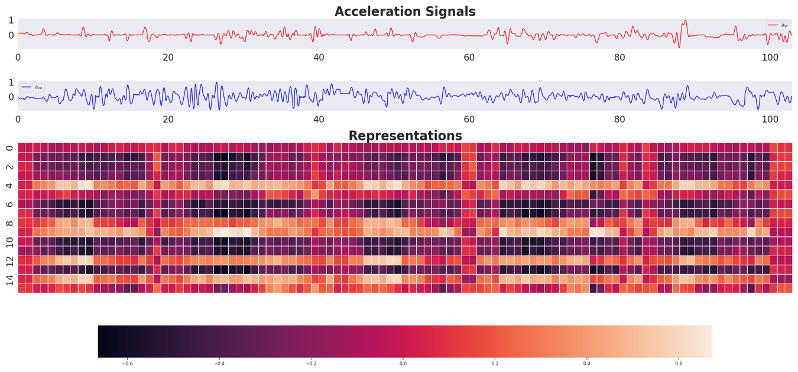}
\caption{\textbf{TNC4Maneuvering:} Encoded representations. \label{fig:tncres}}
  \end{subfigure}%
  \hfill
  \begin{subfigure}{0.5\textwidth}
\includegraphics[width=\linewidth]{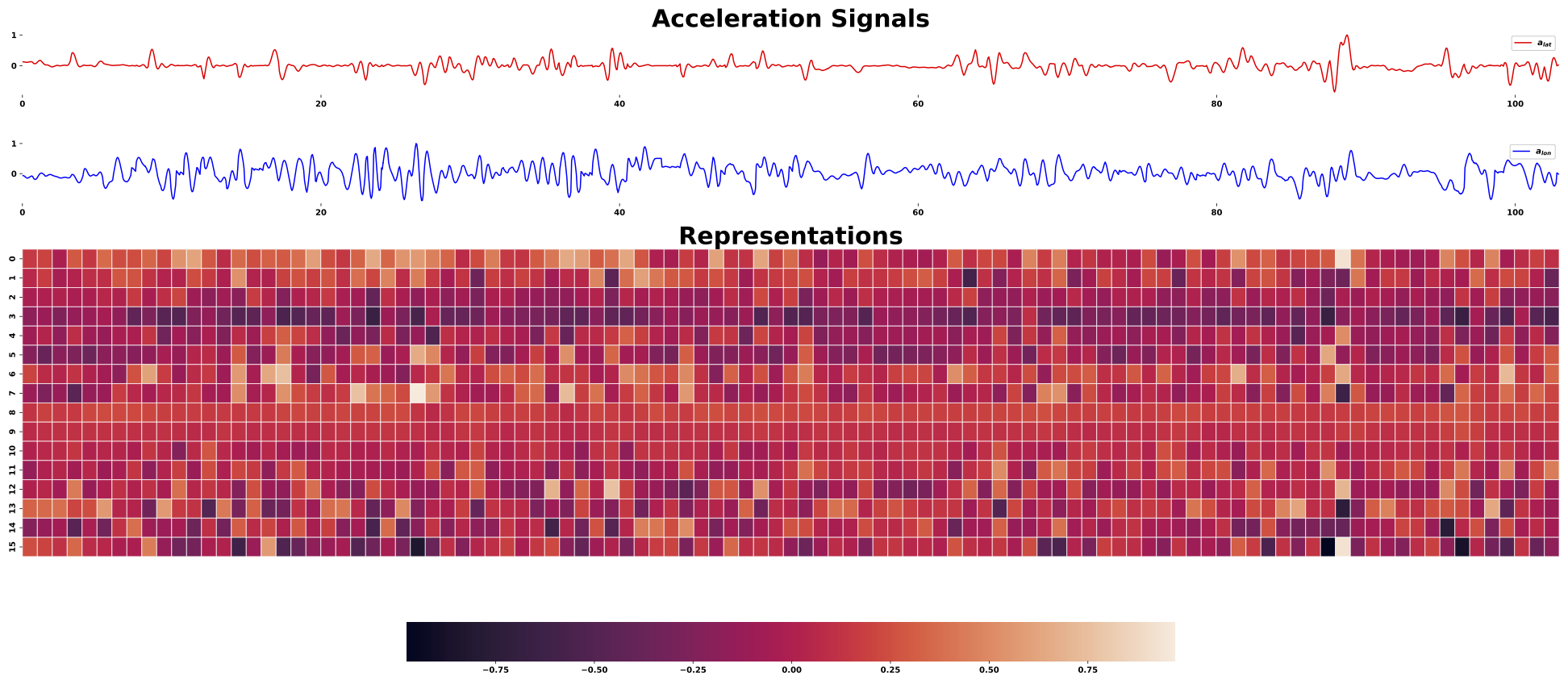}
\caption{\textbf{DLG4Maneuvering:} Locally encoded representations.\label{fig:dglres}}
\end{subfigure}
\vskip 0.1in
\caption{Accelerations and corresponding vector representations~$(M=16)$ encoded using static window $W_t = 19$.
\label{fig:tnc-dgl-res}}
\end{figure*}

\begin{figure*}[t] 
    \centering
    \includegraphics[width=\textwidth]{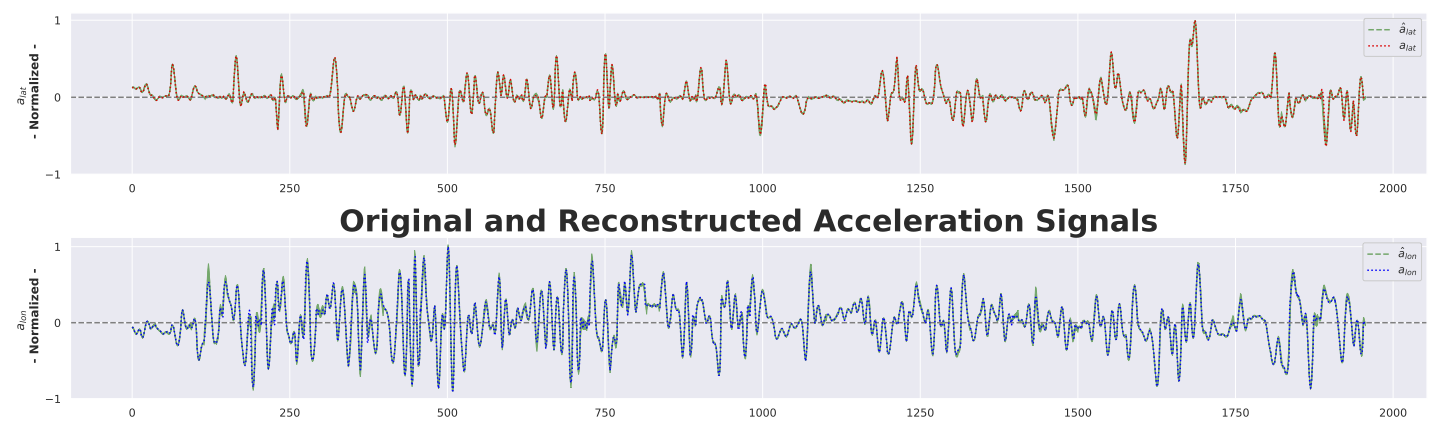}
        \caption{Original $(a_{lat}, a_{lon})$ and reconstructions $(\hat{a}_{lon}, \hat{a}_{lat})$ with error bars $(\pm \sigma_{lon}, \pm \sigma_{lat})$ of bivariate signals using DLG4Maneuvering with a static window size of $W_t = 19$.}
    \label{fig:reps_vis}
\end{figure*}

TNC4Maneuvering, an extension of \cite{tlebese}, is implemented in the PyTorch framework (v1.12.1). On the other hand, DLG4Maneuvering, our extended adaptation of \cite{tonekaboni2022decoupling, otten2021event}, is implemented in the Tensorflow framework (2.6.2). All experiments are conducted using a single Nvidia Tesla P40 GPU with CUDA 11.2.152.\\

In our dataset, we apply only normalization as a pre-processing stage to avoid statistical biases that could lead to misinterpretation of the encoded results. This approach differs from the works of \cite{sajal2019cost}, where (Debauches) wavelet filtering was applied to remove high-frequency noise in signals via denoising and in \cite{shouno2018deep} where the original MTS were down sampled from 40 Hz to 20 Hz.\\

Vehicle maneuvering, a central automotive problem for understanding driving behavior from sensory signals, is explored using a Peugeot 208 model, serving as a fleet car. The operation time accumulates as the duration during which driving activities are collected by various sensors. This work focuses specifically on two accelerations: the lateral acceleration $(a_{lat})$, an effective measure of cornering (negative for right turns, $0$ for straight lines or braking, and positive for left turns), and the longitudinal acceleration $(a_{lon})$, representing straight-line acceleration (negative for braking, $0$ for constant speed, and positive for acceleration). Both accelerations are reported as fractions of gravitational acceleration (ms$^{-2}$). The inputs are normalized such that each $X_i = x_i/x_{\max} \in [-1, 1]$, where $x_{\max} = \max|x_i|, i = \{1, 2\}$, preserving zero values on each feature. Here, we consider only one bivariate sample signal with a signal length of 1957, covering a time of 584 minutes and a mileage of 20 kilometers as depicted in Figure~\ref{fig:reps_vis} (excluding the reconstruction parts in green color).\\

Since there is no prior domain knowledge on maneuver states, we propose a statistical approach serving as ground truth labels to which is different from the works of authors \cite{sarker2021driving}. We additional add a label column with four maneuvering activities, namely state 0: both $a_{lat} \text{ and } a_{lon}$ are stationary, state 1: only $a_{lon}$ is stationary, state 2: only $a_{lat}$ is stationary, and state 3: both $a_{lat}$ and $a_{lon}$ are non-stationary. Stationarity refers to cases where the ADF (p-values $> 0.01$) for each window-size of $250$ of signals as an additional column. These states are treated as ground truth without loss of generality. 

\subsection{Results Discussion}\label{sec:res_dis}

We provide a detailed comparative interpretation of the results obtained from the two methods. The quantitative evaluations are presented in Table~\ref{table:states}, while the visualized evaluations of representations are shown in Figure~\ref{fig:tnc-dgl-res}, and the reconstructed signals are displayed in Figure~\ref{fig:reps_vis}.\\

Based on the results presented in Table \ref{table:states}, DGL4Maneuvering consistently outperforms TNC4Maneuvering across multiple downstream tasks. \\

In the classification tasks, DGL4Maneuvering achieves significantly higher AUPCR (Area Under the Precision-Recall curve) and accuracy values compared to TNC4Maneuvering. Specifically, DGL4Maneuvering has AUPCR of 0.998 and an accuracy of 99.70, while TNC4Maneuvering lags behind with an AUPCR of 0.529 and an accuracy of 53.31. This suggests that DGL4Maneuvering is more effective in correctly classifying maneuvering states, demonstrating its superior performance in tasks requiring precise classification such as driving behavior. Another takeaway which supports the claim that global representations are superior at capturing driving behavior, also the high scores can be attributed to the fact that DGL4Maneuvering can identify samples of similar behavior with ease better using global representations which is regardless of the changes in time which TNC4Maneuvering does not have such a component.\\

For clustering tasks, TNC4Maneuvering shows an advantage in silhouette score over DGL4maneuvering. The Silhouette score measures the similarity of an object to its own cluster compared to other clusters. TNC4Maneuvering achieves a silhouette score of 0.273 compared to DGL4maneuvering with a score of 0.143, indicating a better-defined and well-separated clustering structure. Whereas DGL4maneuvering for is more impressive than TNC4Maneuvering. Overall, both the scores are not as impressive as we desire them to be.\\

For regression tasks, both models exhibit negative values for $R^2$ (coefficient of determination), suggesting challenges in predicting the variability of the response data around its mean. However, DGL4Maneuvering outperforms TNC4Maneuvering with a higher negative $R^2$ value of -0.062 compared to TNC4Maneuvering -0.343. This indicates that DGL4Maneuvering provides a relatively better fit to the liner regression. Regarding the loss values, DGL4Maneuvering achieves a lower loss of 0.355 compared to TNC4Maneuvering 0.449, further emphasizing its superior performance. Since the Linear regression performs the least consistently well, this indicates that localized manually extracted maneuver behaviors are not linearly explained by representations from both methods. A perspective would be to resort to a non-linear mapping to better link the proposed representations with the quantity interest or further improve the quality of the representations. Overall, DGL4Maneuvering consistently demonstrates superior performance across various downstream tasks, making it the preferred choice over TNC4Maneuvering in this comparative analysis.\\

Figure~\ref{fig:tnc-dgl-res} depicts representations of both methods that are obtained from learning bivariate accelerations encoded with a static window-size $W_t = 19$ into a vector representations of size $16$. Figure \ref{fig:tncres} shows both accelerations and their learned representations from TNC4Maneuvering, in this case we can see that both accelerations ($a_{lon}, a_{lat}$) tend to have simultaneous activities, it can also be observed with correspondence to the color code in the representation space that are similar to when there is low activity. Overall, it appears that $a_{lon}$ strongly influences the characteristics of the representations. This is due to the vehicle executing less full turns and making accelerations and deceleration more on this particular dataset.\\

While Figure \ref{fig:dglres} also shows both accelerations and the learned representations using DLG4Maneuvering, although both accelerations ($a_{lon}, a_{lat}$) tend to have simultaneous activities, it is not visually trivial to observe the correspondence of these activities on the representations to the color code in the representation space. Therefore, the similarities of when there is low activity and high activity are not trivially observable, this can be due to the fact that post encoding, there is a post processing step where the outputs of the local encoder are an input to the various kernels (RBF and Matern32) before they are a final representation output which is by design.\\

Overall, the representations show that TNC4Maneuvering has superior representations that enable easy interpretation compared to those from DGL4Maneuvering. Secondly, it can be noted that both representation dimension sizes are large as there is some repetition like behavior in their activity and some of the dimensions seem to be noisy and less informative, further indicating the need and importance of optimizing the representation space.\\

Depicted in Figure~\ref{fig:reps_vis} is the reconstruction of signals from the DLG4Maneuvering generative model. Overall, both $(\Hat{a}_{lat}, \Hat{a}_{lon})$ serve as adequate reconstructions of the original input signals $(a_{lat}, a_{lon})$. They coincide with the original signals and fall within the defined error bars of $(\pm \sigma_{lat}, \pm \sigma_{lon})$. Therefore, this supports the notion that the generator is competent at reconstructing the overall signals, capturing both the moving average and the min-max parts of the signals where most values perfectly coincide in both signals. This trait can be attributed to the quality of both local and global representations, in this case.\\

On the other hand, advantages of this reconstruction is that if it were deployed for further subsequent task such as anomaly detection of anomalous driving behaviors and forecasting of average driver behaviors even in the presence of missing values, it would still out-perform most methods. On the other hand, as much as the representations from DLG4Maneuvering in Figure~\ref{fig:dglres} are not as best as those from TNC4Maneuvering in Figure~\ref{fig:tncres}, we can see that at least they are useful enough to give an adequate reconstruction of original signals therefore proving the importance of getting interpretable representations and perfect reconstructions. This approach further enables clearer understanding of the severity or gentleness of driver maneuvers. Our reliance on the method ability to produce perfect reconstruction samples from interpretable local and global representations is a critical evaluation criterion.

\begin{figure*}[t]
    \centering
    \includegraphics[height = 0.4\textwidth, width=.8\textwidth]{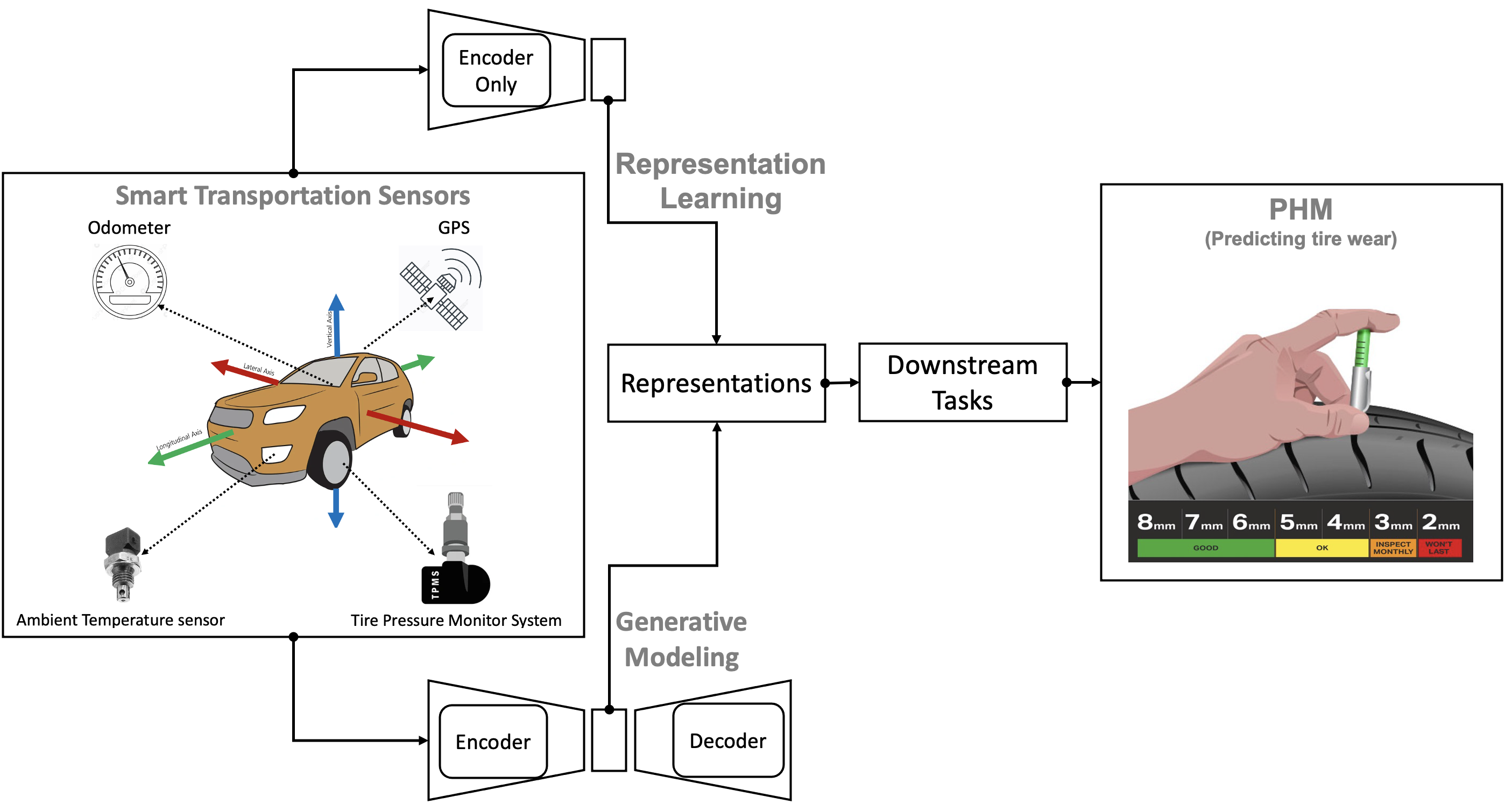}
        \caption{Global objectives and workflow.\label{fig:thesis_goals}}
\end{figure*}

\section{\uppercase{EXPECTED OUTCOME}}
DLG4Maneuvering excels in three comparable downstream tasks and introduces additional reconstructions for input Multivariate Time Series (MTS). However, it falls short in generating interpretable representations compared to TNC4Maneuvering. In our concurrent work, we devised an optimal window selection algorithm and methods for determining the representation size. We attribute the inferior performance of TNC4Maneuvering to a sub-optimal and smaller window size.\\

The outlined goals and overarching outcomes of this work in Figure \ref{fig:thesis_goals}, read from left-to-right, illustrate the achieved milestones, emphasizing the interconnected nature of the goals. The remaining aspect in this work involves leveraging the entire 2.5 years of data for accurate tire wear predictions. Future efforts include scaling both methods to ensure their suitability for other downstream tasks, particularly as meaningful Prognostics and Health Management (PHM) tasks like predicting tire wear. This will be accomplished by utilizing representations from a superior model of the two, one capable of performing well regardless of Multivariate Time Series (MTS) length and complexities.\\

Regarding the scalability issue, the complexities associated with the entire 2.5 years dataset pose significant challenges. It would require weeks to months, along with additional resources, including an increased number of GPUs and memory, to train, test, and evaluate on current university provided environment setup. Currently, TNC4Maneuvering and DLG4Maneuvering take approximately 19 hours and 1.5 days, respectively, for 30 epochs of training. Second, the used data subset in Figure~\ref{fig:reps_vis} constitutes only $0.0695\%$ of the entire dataset, which totals $2813851$. Third, our reliance on the shared university cluster is constrained to jobs that take no more than 7 days, involve 2 GPUs, and use 64GB of CPU memory, regardless of the task. Hence, we are exploring High-Performance Computing (HPC) methods, such as parallelization and distributed training, to assess the feasibility of leveraging the entire dataset within the current environment. Achieving success in this sub-task would allow the realization of our global objectives depicted in Figure \ref{fig:thesis_goals}.

\section{\uppercase{STAGE OF THE RESEARCH}}
This research primarily focuses on developing machine learning methodologies to extract actionable insights, specifically driving behavior as representations, from complex vehicle time series datasets, such as acceleration signals. While these datasets offer rich information about individual driving behavior, their complexity presents challenges that hinder their effective use in real-world automotive industry settings. Our work addresses these challenges through novel approaches that uncover and understand the underlying factors in time series data, particularly in settings with various interdependent and non-continuous labels, rendering on-shelf supervised methods useless.\\

This work has successfully explored and applied deep learning solutions for proactive Prognostics and Health Management (PHM) in smart mobility using vehicle datasets. Our primary objective is to bring advanced methods, particularly deep learning models, closer to adoption in the automotive industry. Current achievements include identifying interpretable representations, deploying them in subsequent machine learning tasks, and using quantitative and visualizable metrics for predicting tire wear. However, our approaches face challenges in scaling to handle the vast amount of available data. As of writing this paper, the author is in the first half of the third and final year of the doctoral studies. Moving forward, the focus will be on exploring scalability strategies to ensure compatibility with computational resource limitations while maintaining effective performance.

\section*{\uppercase{Acknowledgments}}
I am grateful for the support and funding provided by the European Union's Horizon 2020 research program under the Marie Skłodowska Curie project GREYDIENT (Grant Agreement No. 955393). I also extend my sincere thanks to my academic advisors, Cécile Mattrand (Université Clermont Auvergne, Institut Pascal), David Clair (Université Clermont Auvergne, Institut Pascal), Jean-Marc Bourinet (Université Clermont Auvergne, LIMOS) and François Deheeger (MICHELIN) and my colleague at large from MICHELIN R\&D. Additionally, I appreciate the continued support and car dataset provision from Manufacture Française des Pneumatiques Michelin.

\bibliographystyle{apalike}
{\small
\bibliography{example}}
\end{document}